\documentclass{optica-article}

\journal{opticajournal} 

\articletype{Research Article}

\usepackage{lineno}
\usepackage{soul}

\begin{document}

\title{From Fog Chamber to Aircraft Window: Pixel-Registered Imaging and Synthetic Fine-Tuning Enable Cross-Domain Defogging}

\author{Alexander Ingold,\authormark{1} Sabina D. Menon,\authormark{2} Manya Yellepeddy,\authormark{2} Alec Ikei,\authormark{2} John D. Hodges,\authormark{2} Jordan Baker,\authormark{2} Syed N. Qadri,\authormark{2} and Rajesh Menon\authormark{1,*}}

\address{\authormark{1} Department of Electrical and Computer Engineering, University of Utah, Salt Lake City, UT 84112, USA\\
\authormark{2} U.S. Naval Research Laboratory Remote Sensing Division, Washington DC 20375, USA.\\}

\email{\authormark{*}rmenon@eng.utah.edu} 


\begin{abstract*}
A deep defogging pipeline pretrained on controlled laboratory fog
and fine-tuned with domain-randomized synthetic fog applied to
clear outdoor scenes generalizes across a graded sequence of
out-of-distribution settings with no target-domain training, from
chamber-free free-flowing fog to iPhone video recorded through an
aircraft cabin window in flight, an entirely unseen sensor, scene,
and optical path. This directly addresses an open transfer
limitation reported for real-world binocular defogging. Two design
choices support the transfer. First, a single-camera fog imager
photographs a flat-panel display through an artificial-fog
enclosure with a fixed 114~mm scattering path, producing 5{,}495
pixel-aligned foggy/clear pairs. Exact registration permits a
paired Laplacian ratio that predicts per-image restoration quality
far better than single-image proxies (Spearman $\rho = 0.632$
versus $0.399$) and supports pixel-exact $L_1$ reconstruction
training that avoids adversarial hallucination. Second, the
fog-chamber checkpoint is fine-tuned on Mapillary Vistas crops
overlaid with on-the-fly randomized synthetic fog spanning a broad
range of strengths, spatial variations, airlights, and noise
conditions. On a 552-image held-out split, a uniform comparison of
30 restoration backbones places NAFNet at the top
(24.33~dB~/~0.7912~SSIM), with a compact alternative within
1.29~dB at 3\% of the parameter count, and a ResNet-50 classifier
confirms that the restoration preserves semantic content rather
than only pixel-level structure. On unpaired aircraft-window
video, NIQE decreases from a mean of 6.22 to 4.97 after
fine-tuning, with temporally stable output across full-motion
sequences. The same backbone, under
paired supervision, also reaches 20.71~dB~/~0.683~SSIM on a
non-overlapping O-HAZE/NH-HAZE split (a transferability check
rather than a competitive ranking).
\end{abstract*}

\section{Introduction}
\label{sec:intro}
 
Fog and haze attenuate scene radiance along the line of sight and
add scattered ambient light, reducing contrast and suppressing fine
detail. The standard atmospheric scattering model expresses an
observed pixel as a convex mixture of the attenuated scene radiance
and the airlight, weighted by a transmission term that decays with
optical depth \cite{koschmieder1924,narasimhan2002vision}.
Recovering a clear image from a single foggy observation is
therefore underconstrained: scene radiance, airlight, and
depth-dependent transmission are all unknown from one measurement
per pixel \cite{narasimhan2002vision}.
\begin{figure}[htbp]
\centering\includegraphics[width=14cm]{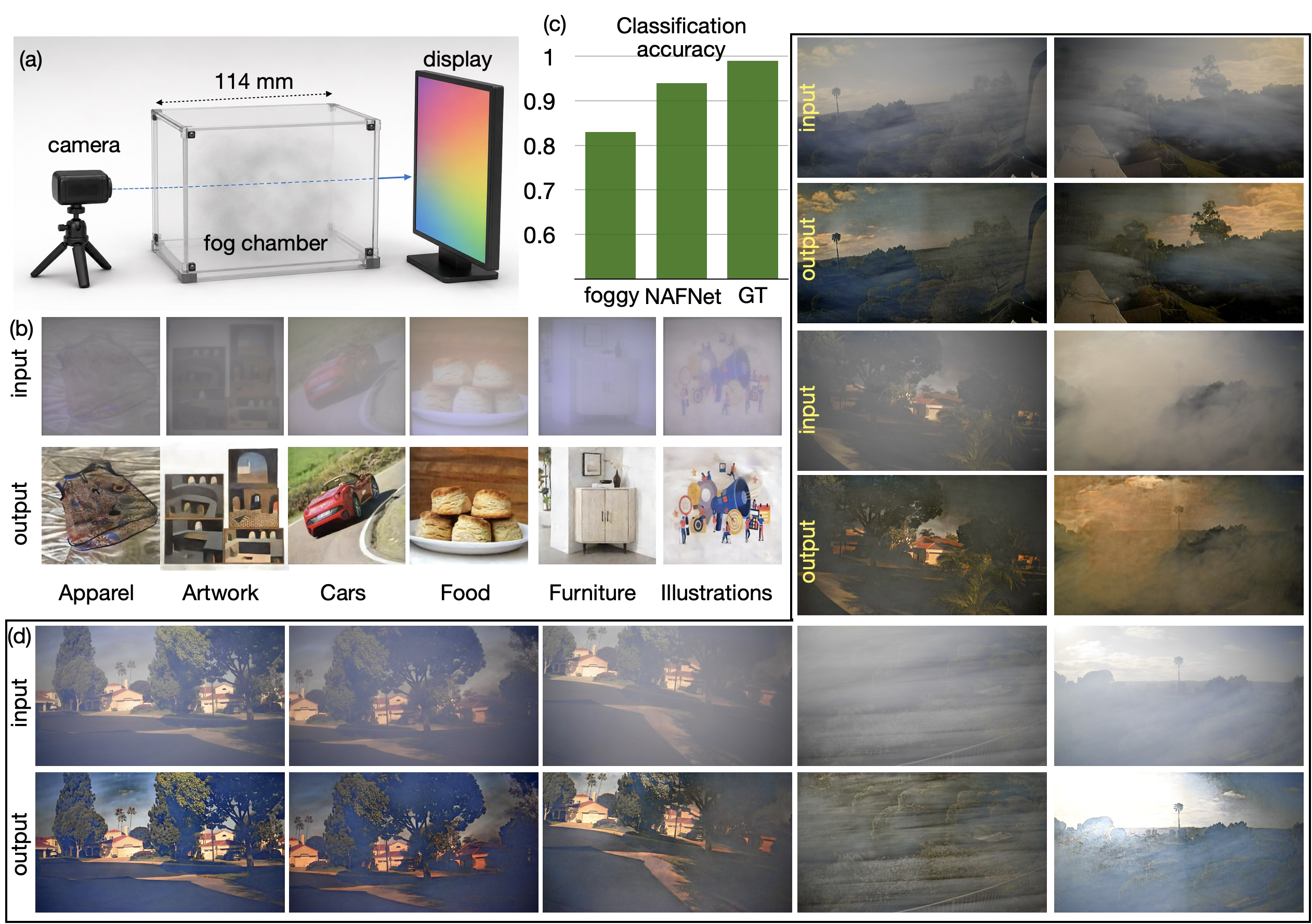}
\caption{Controlled fog-chamber training, benchmark results, and
generalization to chamber-free fog. (a)~Pipeline overview: foggy
images are captured through a fog chamber with a fixed 114~mm
scattering path, and the pixel-registered ground-truth images are
used to train a deep restoration network with pixel-exact $L_1$
supervision. (b)~Representative test-set results across the six
content categories (top row: foggy input; bottom row: NAFNet
prediction); on the 552-image held-out split NAFNet attains mean
PSNR / SSIM of 24.33~dB~/~0.7912. (c)~Category-classification
accuracy of a ResNet-50 evaluated on foggy inputs, NAFNet-defogged
outputs, and ground-truth images, quantifying how much
semantic content the restoration recovers relative to fog
degradation. (d)~Transfer to chamber-free fog generated by the fog
machine placed directly in front of the camera, producing
spatially nonuniform, time-varying scattering across a range of
outdoor scenes (top row: foggy input; bottom row: defogged
output). Together, panels~(b)--(d) show that the trained model
restores both pixel-level structure and semantic content under
controlled fog and continues to remove the bulk of the scattering
when the fog dynamics depart from the chamber regime.}
\label{fig:fog_chamber_overview}
\end{figure}

\subsection{Prior-based and learned restoration}
 
Early methods made the inverse problem tractable with statistical
priors on outdoor scenes, notably the dark channel prior
\cite{he2010single} and the color attenuation prior
\cite{zhu2014single}. These priors are interpretable but fail for
bright or white content, unusual materials, and dense or
spatially uniform fog, and have no mechanism for non-natural
capture geometries. Deep learning shifted the field toward learned
restoration: DehazeNet \cite{cai2016dehazenet} and AOD-Net
\cite{li2017aod} estimated or embedded the scattering model;
GridDehazeNet \cite{liu2019griddehazenet}, FFA-Net
\cite{qin2020ffa}, and DehazeFormer \cite{song2023dehazeformer}
added multi-scale and attention mechanisms; and general-purpose
backbones such as NAFNet \cite{chen2022simple} showed that
streamlined direct-reconstruction networks rival more complex
designs. Surveys nevertheless emphasize a persistent gap between
synthetic-benchmark and real-haze performance
\cite{gui2021comprehensive}, and no large, uniformly trained
benchmark of modern backbones exists for an optically controlled,
exactly paired fog dataset.
 
\subsection{Real-world paired data and the transfer problem}
 
Because aligned foggy/clear pairs are hard to acquire outdoors, the
field has relied on synthetic data, from SYNTHIA
\cite{ros2016synthia} and Foggy Cityscapes
\cite{sakaridis2018semantic} to curriculum adaptation
\cite{sakaridis2018model}. A central limitation is synthetic training can fail when the generated fog distribution is too narrow or does not match real imaging conditions, and unpaired adversarial translation \cite{isola2017pix2pix,zhu2017unpaired} introduces hallucination
risk when faithful reconstruction matters. Real paired benchmarks
such as O-HAZE \cite{ancuti2018ohaze} and NH-HAZE
\cite{ancuti2020nhhaze} anchor the NTIRE challenges
\cite{ancuti2021ntire,histofusionnet2026} but contain only tens of
scenes, and recent methods continue to push real-image dehazing
\cite{chen2024pttd,chen2025dehazexl}.
 
Most directly relevant to the present work, Pollak and Menon
introduced Stereofog, a 10{,}067-pair real-world binocular dataset,
and demonstrated effective \texttt{pix2pix} defogging even under
dense fog \cite{pollak2024stereofog}. That study reached two
conclusions that motivate this paper. First, models trained on
generic synthetic fog failed on real images. Second, and more
fundamentally, the learned models transferred poorly to different
cameras and scenes without retraining, which the authors attributed
to adaptation to specific recording hardware. The binocular design
also captures the two views with separate cameras, introducing
residual misregistration that required shift-robust metrics and
favored adversarial translation, whose hallucination tendency the
authors flagged as a limitation. Whether a controlled,
exactly registered capture system combined with randomized synthetic fog fine tuning can instead produce models that transfer across hardware and scenes has remained open.
 
\subsection{Gaps and contributions}
\label{sec:contributions}

The prior literature leaves four gaps: no optically controlled,
exactly paired fog dataset large enough for a uniform many-model
benchmark; little use of paired data to characterize the fog
itself and cross-domain generalization almost never tested from a
controlled imager to a completely different consumer sensor and
scene, let alone to motion video. We address these with an
end-to-end study centered on data, a paired difficulty metric, and
a domain-randomized synthetic pipeline; we deliberately use existing
restoration backbones rather than introducing a new architecture,
so improvements isolate to the data and training regime. Our
contributions are:

\begin{enumerate}

\item \textbf{Cross-domain transfer across a graded out-of-distribution
sequence, including aircraft-window iPhone video.} A model trained
only on controlled fog-chamber pairs and domain-randomized
synthetic outdoor fog defogs (i)~chamber-free free-flowing fog
generated by a fog machine placed in front of the camera, in
which the optical depth along the line of sight is no longer fixed
by a geometry (Fig.~\ref{fig:fog_chamber_overview}d), and (ii)~iPhone
video captured through an aircraft cabin window in flight, with
no target-domain training and temporally stable output across
full-motion sequences (Fig.~\ref{fig:aircraft}; Visualizations~1 and~2). This directly addresses the cross-hardware
transfer limitation reported for real-world binocular
defogging~\cite{pollak2024stereofog}.

\item \textbf{A controlled, exactly registered fog imager and a
5{,}495-pair dataset.} A display-based imager photographs a flat-panel display through an artificial-fog enclosure
with a fixed 114~mm scattering path. Unlike binocular
capture~\cite{pollak2024stereofog}, single-camera display imaging
gives exact pixel registration, which both removes the need for
shift-robust metrics and enables direct reconstruction training
that avoids adversarial hallucination.

\item \textbf{A paired fog-difficulty metric.} The mean-absolute
Laplacian ratio, a paired refinement of the single-image Laplacian
variance used to grade fog density in prior
work~\cite{pollak2024stereofog}, predicts per-image restoration
quality with Spearman $\rho = 0.632$ (category-centered $0.590$),
far exceeding the best single-image proxy ($\rho = 0.399$).

\item \textbf{A 30-model benchmark with a semantic-content check.}
Thirty restoration backbones are trained under identical conditions
on the fog-chamber dataset, establishing a reproducible baseline
across contemporary defogging architectures and generic
image-to-image models. NAFNet~\cite{chen2022simple} ranks highest
(24.33~dB~/~0.7912~SSIM), while SpecAT~S2~\cite{yao2024specat} is
within 1.29~dB at 3\% of the parameter count. A ResNet-50
classifier evaluated on foggy, defogged, and clear inputs shows
that classification is already feasible directly from foggy
images, and that NAFNet defogging closes most of the residual gap
to ground truth (Fig.~\ref{fig:fog_chamber_overview}c), confirming
that the restoration preserves semantic content rather than only
low-level appearance.

\item \textbf{Domain-randomized synthetic fine-tuning and
task-specific checks.} The fog-chamber NAFNet checkpoint is
fine-tuned on Mapillary Vistas clear crops with on-the-fly spatial
synthetic fog. Light-fog measurements set only loose generator
ranges; the final transfer model uses randomized fog strength,
airlight, spatial variation, contrast, color, and noise rather than
optimization to a narrow unpaired reference set. The same backbone
is also evaluated under paired task-specific fine-tuning on O-HAZE,
NH-HAZE, and the NTIRE~2026 nighttime probe.

\end{enumerate}
 
The dataset, code, simulator, trained models, and demonstration videos are released
publicly with the reproducibility package. Section~2 describes the chamber geometry and paired metric,
Section~3 the methods, Section~4 the results, Section~5 the
discussion, and Section~6 concludes. The supplement provides split audits, metric definitions, the full benchmark table, classification and dark-channel-prior checks, simulator details, public real-haze protocols, fog-density statistics, and paired structure-loss analysis.

\section{Chamber geometry and Paired Metric}
\label{sec:fogphysics}
 
\subsection{Scattering model background and chamber geometry}
 
The atmospheric scattering model \cite{narasimhan2002vision} writes
the observed pixel as
 
\begin{equation}
  \mathbf{I}(\mathbf{x}) =
    \mathbf{J}(\mathbf{x})\,T(\mathbf{x})
    + \mathbf{A}\bigl[1 - T(\mathbf{x})\bigr],
  \qquad
  T(\mathbf{x}) = e^{-\beta\, d(\mathbf{x})},
  \label{eq:scattering}
\end{equation}
 
\noindent where $\mathbf{J}$ is the clear radiance, $\mathbf{A}$ the
global airlight, $T$ the transmission, $\beta$ the extinction
coefficient, and $d$ the path length. For the fog imager $d$ is
fixed by the chamber geometry. We use this model as a physical motivation for the synthetic fog simulator, not as a radiometric inversion of the fog-chamber data. The display, camera response, exposure and fog radiance were not independently calibrated, so we do not estimate a physical transmission map or extinction coefficient from the image pairs. The acrylic fog chamber had a 114~mm fog path-length and 133 $\times$ 114~mm transverse dimensions (see Fig.~\ref{fig:fog_chamber_overview}). Artificial
fog was generated by an Eliminator Lighting VF400-E theatrical fog
machine using propylene-glycol fluid.
 
\subsection{A paired fog-difficulty metric}
\label{subsec:fogdifficulty}
 
Prior real-world work graded image-level fog severity by the variance of the
Laplacian of the foggy image alone \cite{pollak2024stereofog}.
Exact registration lets us instead form a \emph{paired}
high-frequency retention ratio,
 
\begin{equation}
  R_{\nabla^2} =
    \frac{\mathrm{mean}\!\left(|\nabla^2 \mathbf{I}_{\mathrm{fog}}|\right)}
         {\mathrm{mean}\!\left(|\nabla^2 \mathbf{I}_{\mathrm{clear}}|\right)},
  \label{eq:laplacian_ratio}
\end{equation}
 
\noindent which measures how much of the clear image's local
high-frequency structure survives the fog and normalizes out scene
content. On the 552-image split, $R_{\nabla^2}$ at one-pixel scale
predicts per-image NAFNet PSNR with Spearman $\rho = 0.632$
(category-centered $0.590$), substantially better than the best
single-image proxy (dark-channel 90th percentile, $\rho = 0.399$;
category-centered $0.312$). Supplementary Fig. S7 summarizes the metric ranking, per-image relationship to NAFNet PSNR, and a representative paired Laplacian example. The full-metric search is in Supplement section S10 and the image-domain power-spectral-density (PSD) analysis is in Supplement section S9. The PSD analysis shows that high-frequency power retention was 8.9$\times$ lower than low-frequency retention in paired fog/clear images. 
 
\section{Methods}
\label{sec:methods}
 
\subsection{Apparatus and dataset}
\label{subsec:apparatus}
 
As illustrated in Fig.~\ref{fig:fog_chamber_overview}(b), a flat-panel LCD presents the ground-truth target through an acrylic
enclosure of internal path length $114~\mathrm{mm}$ and transverse
cross-section $133 \times 114~\mathrm{mm}$, with fog from the VF400-E
machine; in some runs a vertical linear polarizer was placed before
the camera (noted where used). Single-camera display imaging yields
exact pixel registration without post-hoc alignment, in contrast to
binocular capture \cite{pollak2024stereofog}, and lets us train with
pixel-exact L1 reconstruction rather than adversarial losses (Fig.~\ref{fig:fog_chamber_overview}a).
Ground-truth images are drawn from a public $130\mathrm{k}$-image
archive at $512 \times 512$ across six categories \cite{singh130kimages}.
A deterministic every-tenth split yields 4{,}943 training and 552
held-out pairs (92 per category; 5{,}495 total). Full dataset roles
and split audits are in Supplement section S1.
 
\subsection{Benchmark, classification, simulator, and adaptation}
\label{sec:methods-benchmark}
We trained thirty restoration backbones on the fog-chamber training
split under identical conditions: pixel-exact $L_1$ loss, Adam
optimizer~\cite{kingma2015adam}, learning rate $10^{-4}$, 50 epochs,
and $512\times512$ inputs (memory-heavy models used crops at the
same resolution). Each model was evaluated on the 552-image held-out
split by mean MAE, MSE, PSNR, and SSIM~\cite{wang2004ssim}, with
LPIPS~\cite{zhang2018lpips} additionally reported for the top five.
NAFNet~\cite{chen2022simple} ranked highest and was used as the
backbone for every downstream fine-tuning stage in this paper except for the ablation run with random weight initialization.

To assess whether restoration preserves semantic content rather than
merely pixel-level structure, we trained a ResNet-50
classifier~\cite{he2016deep} on five dataset categories excluding illustrations and
evaluated it on clear, foggy, and NAFNet-defogged images
(Fig.~\ref{fig:fog_chamber_overview}c). Classification from foggy
inputs already retains most of the ground-truth accuracy, and
NAFNet-defogged inputs close most of the remaining gap, indicating
that the restoration recovers semantically discriminative content
rather than only low-level appearance.

To bridge from chamber fog to outdoor fog, the fog-chamber NAFNet
checkpoint was fine-tuned on clear Mapillary Vistas
crops~\cite{neuhold2017mapillary} with on-the-fly spatial synthetic
fog (Supplement section S6). The split contained 18{,}000 training,
2{,}000 validation, and 5{,}000 test clear images. Training used
$512\times512$ crops, AdamW~\cite{loshchilov2019adamw}, cosine
learning-rate decay from $10^{-4}$ to 0, and one epoch capped at
2{,}600 training batches. Each crop sampled fog strength, spatial
variation, airlight color, contrast, saturation, edge veil, blur,
and noise. Free-space light-fog image statistics were used only to
choose plausible initial ranges for these random variables, not to
fit a fixed generator and not as training, validation, or test
images. Synthetic validation and test PSNR were 21.38~dB and
21.17~dB, respectively, and are treated only as synthetic-training
diagnostics. The matched no-pretraining ablation reached
16.47~dB validation and 16.58~dB test PSNR under the same synthetic
recipe.

For paired real-haze adaptation, the fog-chamber NAFNet was
fine-tuned with paired supervision on a non-overlapping mixed split
of O-HAZE~\cite{ancuti2018ohaze} and NH-HAZE~\cite{ancuti2020nhhaze}.
We additionally report a two-image internal-test probe of the NTIRE
2026 nighttime dehazing
challenge~\cite{ancuti2026ntire,ancuti2026nt} as a sanity check
of paired adaptation under nighttime conditions; no challenge rank
is claimed. Splits, hyperparameters, and per-image metrics are
detailed in the Supplement.

Complete protocols for the benchmark, randomized synthetic fine-tuning, public
paired-haze checks, NTIRE probe, fog statistics, PSD analysis, and
paired structure-loss analysis are detailed in Supplement sections
S3--S10.
 
\subsection{Out-of-distribution evaluation}
\label{sec:ood}

We applied the synthetic fine-tuned model, with no target-domain
training, to two evaluations that progressively stress its
generalization: outdoor captures with a fog machine placed
directly in front of the camera (no enclosure), and video from a
consumer iPhone (iPhone~13, iOS~26.5) recorded through an aircraft
passenger-cabin window in daylight and at dusk.

The chamber-free fog captures probe transfer to spatially
nonuniform, time-varying scattering, in which the optical depth
along the line of sight is no longer fixed by a geometry as it is
in the chamber. Representative frames are shown in
Fig.~\ref{fig:fog_chamber_overview}d; the model removes the bulk of the
scattering while preserving scene structure, indicating that the
learned restoration is not tied to the chamber's fixed 114~mm
path length.

\begin{figure}[htbp]
\centering\includegraphics[width=14cm]{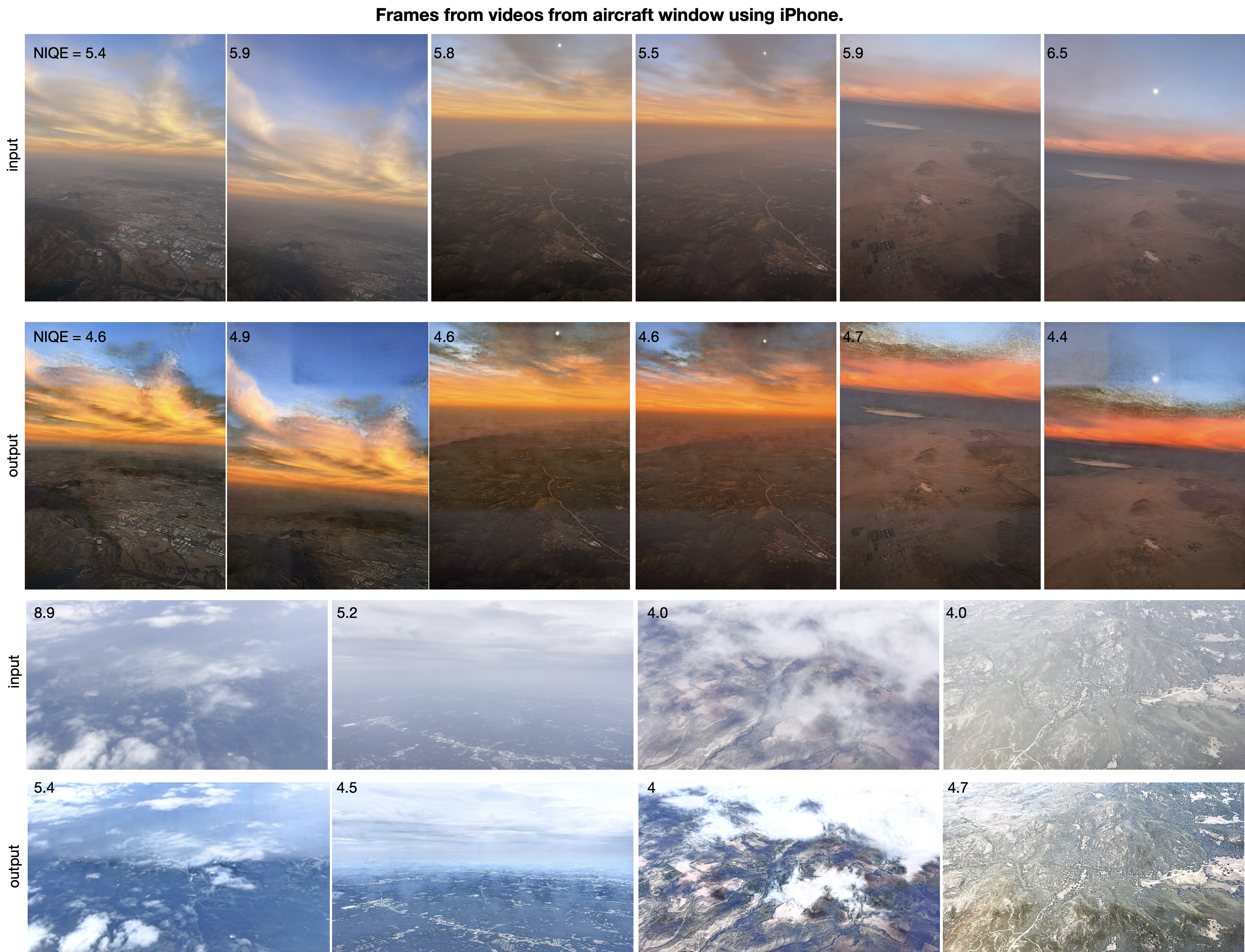}
\caption{Testing our trained model on videos captured using an iPhone camera through an aircraft window at high altitudes under different conditions including dusk and bright daylight. In all cases, the trained network is able to reduce visible fog, but loss of dynamic range especially at the brightest pixels are seen. This is particularly impressive since such images and camera are completely foreign to the network. Since ground-truth is not available for these images, we report the NIQE metric in each image. Also see Visualizations 1 and 2.}
\label{fig:aircraft}
\end{figure}

The aircraft-window evaluation is the strongest transfer test in
the paper. Four factors make it out of distribution simultaneously.
First, the cabin glass introduces scattering and specular
reflections that are absent from training. Second, the aerial scene
content (sky, clouds, terrain at oblique view) is unlike any image
in the training set. Third, the iPhone sensor and its on-device
tone mapping differ from the chamber camera. Fourth, video demands
frame-to-frame temporal consistency that single-frame training
never supervised. With no aligned references we cannot report
PSNR/SSIM; instead we report the no-reference Natural Image Quality Evaluator (NIQE)~\cite{mittal2013niqe}, for which lower values indicate higher
perceived quality, together with the demonstration videos
(Visualizations~1 and~2). Representative frames are shown in
Fig.~\ref{fig:aircraft} with the per-frame NIQE value overlaid on
each panel.

\section{Results}
\label{sec:results}
 
\subsection{Zero-shot transfer to aircraft-window video}
\label{subsec:transfer}
 
The pipeline's central result is cross-domain transfer with no
target-domain training. Fine-tuning the fog-chamber NAFNet on synthetic outdoor fog reached 21.17~dB on a held-out split of the same synthetic distribution; this number quantifies fit to the simulator, not real-fog transfer, which we evaluate below. The decisive
evidence is qualitative transfer to real, unrelated fog
(Fig.~\ref{fig:aircraft}). The fog-chamber model alone
over-corrects outdoor and aircraft-window inputs with color shifts
and artifacts, while randomized synthetic fine-tuning removes
this failure.
 
Applied to iPhone video through an aircraft cabin window, the pipeline produces temporally stable output across full-motion sequences without visible flicker or frame-to-frame color drift in the demonstration clips (Visualizations 1 and 2). While NIQE did not decrease for every image, NIQE decreased on average from 6.22 to 4.97 for the unpaired aircraft-window examples (Fig.~\ref{fig:aircraft}; Supplementary Table S12). This directly answers the open question from real-world binocular defogging, where models did
not generalize across cameras and scenes without retraining
\cite{pollak2024stereofog}: a controlled exactly registered capture
followed by synthetic fine-tuning yields a model
that does. Two demonstration videos accompany the paper and are the
most direct evidence of practical utility; static metrics cannot
establish temporal stability.

 \begin{figure}[htbp]
\centering\includegraphics[width=14cm]{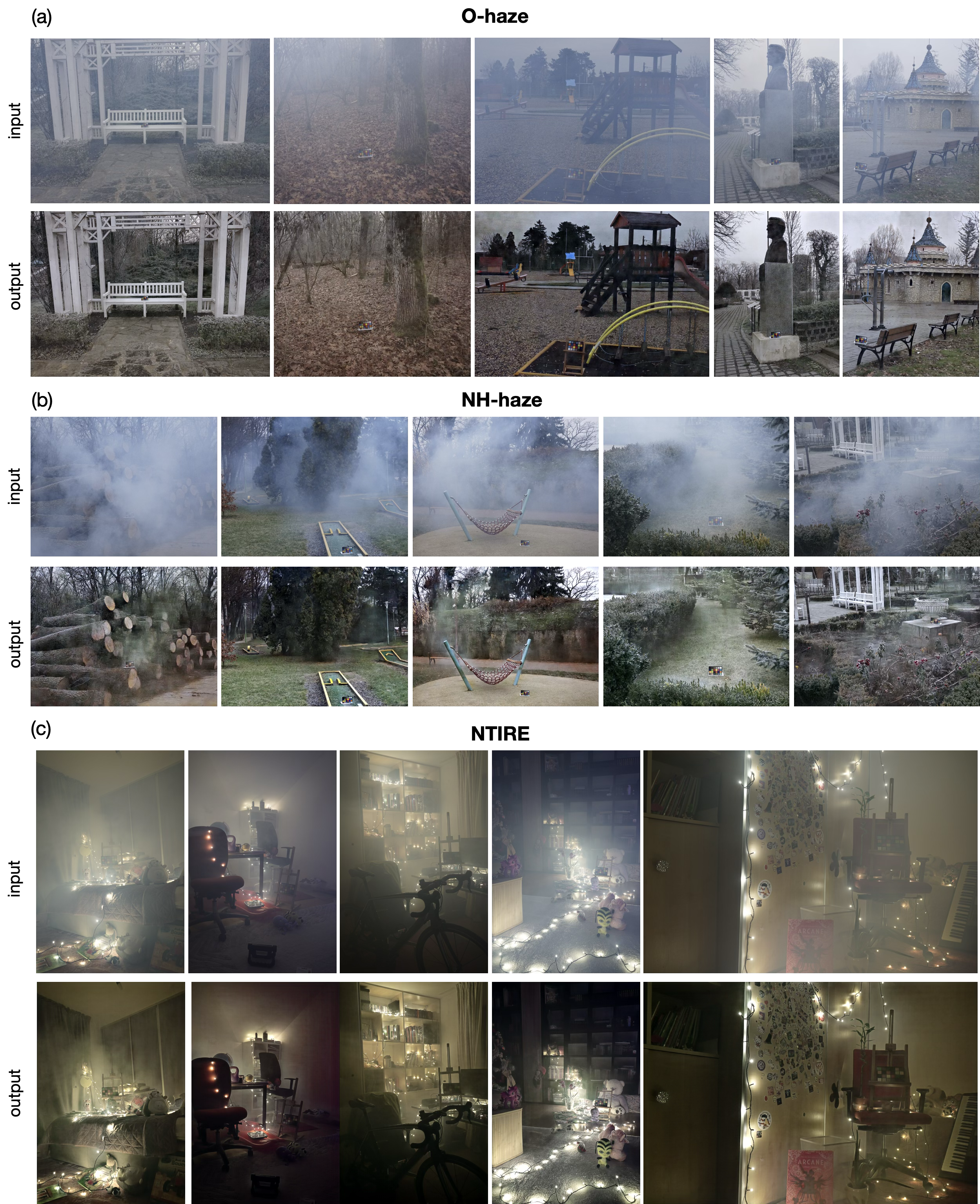}
\caption{Task-specific public-haze and nighttime-haze checks. (a) O-HAZE, (b) NH-HAZE and (c) NTIRE. Various exemplary images with different fog severities are successfully defogged. The previously trained models were fine-tuned to the dataset for best results. Quantitative split definitions and per-image metrics are reported in Supplement section~S7}
\label{fig:O_NH_haze}
\end{figure}
\subsection{Public paired real-haze adaptation}
\label{subsec:realhaze_results}
 
The same backbone adapts under paired supervision. On a
non-overlapping mixed split it reached $20.71~\mathrm{dB}$ /
$0.683$~SSIM (O-HAZE subset $22.65$ / $0.726$; NH-HAZE subset
$18.77$ / $0.640$; the $3.88~\mathrm{dB}$ gap reflects
non-homogeneous haze). Exemplary results are summarized in Fig.~\ref{fig:O_NH_haze}. These are controlled transferability checks,
not competitive claims; dedicated O-HAZE methods exceed
$24~\mathrm{dB}$ \cite{chen2024pttd,chen2025dehazexl}. Applying the
synthetic model directly, without real-haze fine-tuning,
over-corrects these inputs, showing the regimes are complementary. The discrepancy with the aircraft-window result likely reflects the absence of ground truth there: over-correction that depresses PSNR on O-HAZE/NH-HAZE may go partially undetected in a reference-free NIQE evaluation. Real-fog video with aligned references is the natural next-step measurement (see Limitations). 
An NTIRE 2026 nighttime probe reached $24.77~\mathrm{dB}$ /
$0.770$~SSIM on a two-image internal split (a probe only, no
challenge rank claimed)\cite{ancuti2026ntire, ancuti2026nt}. Category recognizability and
classification results are in Supplement~S4.

\subsection{Benchmark and paired difficulty metric}
\label{subsec:benchmark_results}
 
\begin{figure}[htbp]
\centering\includegraphics[width=14cm]{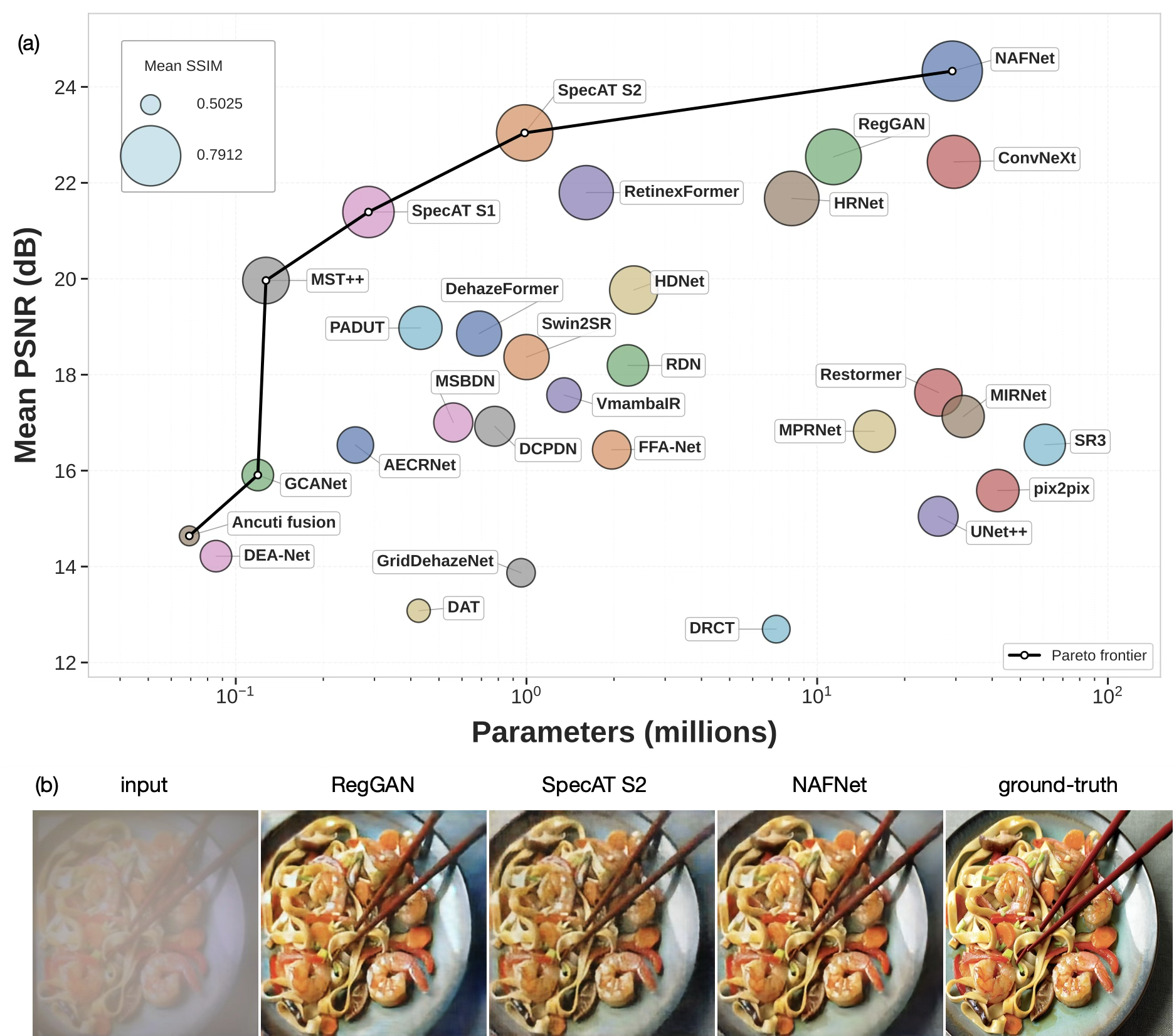}
\caption{Defogging benchmark across established models on the fog-chamber dataset. (a) Thirty-model benchmark summary: marker area encodes mean SSIM, the horizontal axis shows parameter count, and the vertical axis shows mean PSNR; the black line marks the Pareto frontier for higher PSNR with fewer parameters. (b) Exemplary images from left to right: input foggy, outputs from RegGAN, SpeCAT S2 and NAFNet, and ground-truth.}
\label{fig:benchmark}
\end{figure}

NAFNet ranked highest on the 552-image split with mean PSNR
$24.33~\mathrm{dB}$ (95\% bootstrap CI, 24.00--24.65) and SSIM $0.7912$ (95\% bootstrap CI, 0.7808--0.8012), a $12.53~\mathrm{dB}$ gain
over the foggy-input baseline ($11.80~\mathrm{dB}$); SpecAT~S2\cite{yao2024specat} and
RegGAN followed at $23.04$ and $22.54~\mathrm{dB}$. The parameter-efficiency frontier
(Fig.~\ref{fig:benchmark}) has two anchors: NAFNet
($29.16~\mathrm{M}$ params) for accuracy and SpecAT~S2
($0.99~\mathrm{M}$) for efficiency, the latter within
$1.29~\mathrm{dB}$ at $\approx 3\%$ of the parameters; intermediate models are Pareto-dominated (\emph{i.e.}, worse on both axes than at least one of NAFNet or SpecAT S2). Notably, \texttt{pix2pix}, the adversarial
backbone used for binocular real-world defogging
\cite{isola2017pix2pix, pollak2024stereofog}, ranks low here ($15.58~\mathrm{dB}$):
exact registration favors direct L1 reconstruction over adversarial
translation in this regime. The dark channel prior
\cite{he2010single} reduced PSNR below the foggy input ($11.80
\rightarrow 9.82~\mathrm{dB}$), consistent with the display
back-illumination violating its outdoor assumptions. The full
30-model table is in Supplement table S2.
 
Per-category PSNR spans $3.42~\mathrm{dB}$ (artwork $23.02$,
furniture $26.44$). Categories with dense fine structure retain less high-frequency
content at a given fog level and are harder to restore
(Supplement section S10).

\section{Discussion}
\label{sec:discussion}

The results resolve a tension in prior real-world defogging.
Pollak and Menon found that models trained on a large real-world
paired dataset transferred poorly across cameras and scenes, and
that generic synthetic fog failed on real
images~\cite{pollak2024stereofog}. Two design choices in the present
work recover that transfer. First, single-camera display imaging
gives exact pixel registration, which enables pixel-exact
$L_1$ reconstruction and so avoids the adversarial hallucination
that the binocular setting had to tolerate; the same registration
underwrites the paired Laplacian ratio of Section~\ref{sec:methods}.
Second,  randomized synthetic fine-tuning exposes the fog-chamber
model to a broad family of spatially varying outdoor-like
scattering conditions. The two capture regimes
are complementary points in a design space: binocular capture trades
registration for real three-dimensional scene content, while display
capture trades scene realism for exact registration and content
control. The OE-relevant lesson is that, for a fixed scattering
path and controlled illumination, a carefully posed imaging setup
plus a domain-randomized forward model can substitute for outdoor data
collection in a regime where outdoor collection is impractical.

The aircraft-window result and the O-HAZE/NH-HAZE result appear
asymmetric and deserve direct treatment. The synthetic fine-tuned
model transfers to iPhone aircraft-cabin video without target-domain
training, yet the same model over-corrects O-HAZE/NH-HAZE inputs
under reference-based scoring (Section~\ref{sec:results}). The
most plausible reconciliation is that the aircraft evaluation is
reference-free (NIQE only) and so cannot penalize over-correction
the way PSNR/SSIM does on O-HAZE/NH-HAZE. We therefore frame the
aircraft-window result as evidence that the pipeline produces
visually coherent, temporally stable output on an entirely unseen
sensor, scene, and optical path, rather than as a quantitative
generalization claim. Establishing the stronger claim requires
aligned cross-hardware references, which the present pipeline does
not yet provide.

For practitioners, the benchmark gives two actionable findings.
SpecAT~S2 matches NAFNet within 1.29~dB at roughly 3\% of the
parameters, making it the natural choice when compute or memory is
constrained, and NAFNet is the natural choice when accuracy
dominates. More broadly, exact-registration training favors direct
$L_1$ reconstruction over adversarial translation in this regime,
which inverts the methodological preference reported under
misregistered binocular capture~\cite{pollak2024stereofog}. The
paired Laplacian ratio (Spearman $\rho = 0.632$ versus $0.399$ for
the best single-image proxy) gives a usable per-image difficulty
estimator that, unlike single-image Laplacian variance, is robust
to scene content.

\subsection{Limitations}
\label{sec:limitations}
Four limitations should temper the present claims. (i)~The
fog-chamber capture uses flat display content rather than
three-dimensional scenes, which constrains the realism of training
imagery; the simulator stage and the public-haze adaptation
partially, but not fully, compensate. (ii)~The display, camera
response, exposure, and fog radiance were not independently
calibrated, so the chamber data support learned restoration but do
not support radiometric inversion (Section~\ref{sec:methods}).
(iii)~The public real-haze and NTIRE test splits are small, and
the NTIRE 2026 nighttime evaluation is a two-image probe rather
than a competitive ranking. (iv)~The aircraft-window evaluation
is reference-free; we report NIQE and demonstration videos, but
PSNR/SSIM-grade evidence of cross-hardware generalization will
require small aligned real-fog video sets. Radiometric calibration
of the apparatus and the collection of such aligned real-fog video
sets are the natural next steps.
 
\section{Conclusion}
\label{sec:conclusion}

We presented a controlled display-based fog imager whose exact pixel
registration enables a paired fog-difficulty metric and pixel-exact
reconstruction training. Pretraining on 5{,}495 controlled foggy/clear
pairs and fine-tuning on domain-randomized synthetic fog yields a
pipeline that transfers across a graded sequence of out-of-distribution
settings with no target-domain training: from outdoor captures through
chamber-free free-flowing fog, in which the optical depth along the
line of sight is no longer fixed by a geometry, to iPhone video
recorded through an aircraft cabin window in flight, directly
addressing the cross-hardware transfer limitation reported for
real-world binocular defogging~\cite{pollak2024stereofog}. The
aircraft-window evaluation is reference-free, and a small aligned
cross-hardware test set is the natural next measurement.

Supporting evidence comes from four controlled checks. First, NAFNet
leads a uniformly trained 30-model benchmark on the fog-chamber split
at 24.33~dB~/~0.7912~SSIM, with a compact alternative within 1.29~dB
at roughly 3\% of the parameter count, giving practitioners a usable
accuracy--efficiency frontier. Second, a ResNet-50 classifier on
foggy, defogged, and clear inputs shows that the restoration
preserves semantic content rather than only low-level appearance.
Third, the paired Laplacian ratio predicts per-image restoration
difficulty substantially better than single-image proxies
(Spearman $\rho = 0.632$ versus $0.399$), see Supplement S10. Fourth, the same backbone adapts under paired supervision to O-HAZE/NH-HAZE and to an NTIRE~2026
nighttime probe, confirming that the pipeline is not tied to the
chamber distribution.

Taken together, exact registration and synthetic fine-tuning turn a laboratory fog imager into a model whose output
transfers to free-flowing outdoor fog and to real aerial video
without retraining. The planned release of the dataset, code,
simulator, trained models, and demonstration videos is intended to catalyze further
work on transferable defogging, particularly toward aligned real-fog
video benchmarks that the current literature still lacks.

\begin{backmatter}
\bmsection{Funding} A.I. and R.M. acknowledge funding from US Department of Energy grant \#55801063. S.D.M. and M.Y. acknowledges support through the Science and Engineering Apprenticeship Program (SEAP), administered by the American Society for Engineering Education (ASEE) on behalf of the Office of Naval Research, Department of the Navy. J.D.H. acknowledges support through the Naval Research Enterprise Intern Program (NREAP). 

\bmsection{Acknowledgment} The support and resources from the Center for High Performance Computing at the University of Utah are gratefully acknowledged.

\bmsection{Disclosures} The authors declare no conflicts of interest. The authors used Claude (Anthropic) solely for grammar refinement and organizational suggestions. 

\bmsection{Data Availability} Data and code underlying the results are available in the Supplementary Document and at \url{https://github.com/theMenonlab/defogging}. The fog-chamber dataset is released through Kaggle at \url{https://www.kaggle.com/datasets/alingold/fog-chamber}.

\bmsection{Supplemental document}
See Supplement 1 for supporting content including dataset split audits, metric definitions, the full benchmark table, classification and dark-channel-prior checks, synthetic-fog tuning and fit details, public real-haze protocols, fog-density statistics, PSD analysis, and paired structure-loss analysis.

\end{backmatter}

\bibliography{sample}

\end{document}